\documentclass[10pt,twocolumn,letterpaper]{article}

\usepackage{iccv}
\usepackage{times}
\usepackage{epsfig}
\usepackage{graphicx}
\graphicspath{{./}}
\DeclareGraphicsExtensions{.pdf}
\usepackage{amsmath}
\usepackage{amssymb}
\usepackage{bm}
\usepackage{algorithm}
\usepackage{algorithmic}
\usepackage{array}
\usepackage{makecell}
\usepackage{placeins} %用来处理未处理的浮动图像或者表格，使用命令\FloatBarrier
\usepackage{stfloats}
\usepackage{multirow}

\newenvironment{packed_itemize}{
\vspace{-0.15cm}\begin{itemize}
  \setlength{\itemsep}{1pt}
  \setlength{\parskip}{0pt}
  \setlength{\parsep}{0pt}
}{\end{itemize}}

\usepackage[breaklinks=true,colorlinks,bookmarks=false]{hyperref}

\iccvfinalcopy % *** Uncomment this line for the final submission

\def\iccvPaperID{1976} % *** Enter the ICCV Paper ID here

\ificcvfinal\fi
\begin{document}

%%%%%%%%% TITLE
\title{Learning Policies for Adaptive Tracking with Deep Feature Cascades}
% \title{EDPT: Learning Early Decision Policy for Adaptive Deep Visual Tracking}
% Can we stop at this layer? - 
% learning to make early layer sequential decisions policy adaptive real time deep visual tracking
% Complexity-Aware Deep Visual Tracking
% Learning Timely Policy for Complexity-Aware Deep Visual Tracking
% Adaptive Deep Visual Tracking at Constrained Time Cost

\author{Chen Huang \quad \quad Simon Lucey \quad \quad Deva Ramanan\\
Robotics Institute, Carnegie Mellon University\\
{\tt\small chenh2@andrew.cmu.edu} \quad {\tt\small \{slucey,deva\}@cs.cmu.edu}
}

\maketitle
%\thispagestyle{empty}

%%%%%%%%% ABSTRACT
\begin{abstract}
Visual object tracking is a fundamental and time-critical vision task. Recent years have seen many shallow tracking methods based on real-time pixel-based correlation filters, as well as deep methods that have top performance but need a high-end GPU. In this paper, we learn to improve the speed of deep trackers without losing accuracy. Our fundamental insight is to take an adaptive approach, where easy frames are processed with cheap features (such as pixel values), while challenging frames are processed with invariant but expensive deep features. We formulate the adaptive tracking problem as a decision-making process, and learn an agent to decide whether to locate objects with high confidence on an early layer, or continue processing subsequent layers of a network. This significantly reduces the feed-forward cost for easy frames with distinct or slow-moving objects. We train the agent offline in a reinforcement learning fashion, and further demonstrate that learning all deep layers (so as to provide good features for adaptive tracking) can lead to near real-time average tracking speed of 23 fps on a single CPU while achieving state-of-the-art performance. Perhaps most tellingly, our approach provides a {\bf 100X} speedup for almost 50\% of the time, indicating the power of an adaptive approach.
\end{abstract}

%%%%%%%%% BODY TEXT
\section{Introduction}

Visual Object Tracking (VOT) is a fundamental problem in vision. We consider the single object tracking task, where an object is identified in the first video frame and should be tracked in subsequent frames, despite large appearance changes due to object scaling, occlusion, and so on. VOT is the basic building block of many time-critical systems such as video surveillance and autonomous driving. Thus it is important for a visual tracker to meet the strict constraints of time and computational budget, especially on mobile or embedded computing architectures where real-time analysis perception is often required.

Although much progress has been made in the tracking literature, there still exist tremendous challenges in designing a tracker that has both high accuracy and high speed. Real-time tracking methods like TLD~\cite{6104061} and correlation filters~\cite{henriques2015} usually rely on low-level features that are not descriptive enough to disambiguate target and background. Several recent works~\cite{Danelljan2016,MaICCV2015,Danelljan15,Qi_2016_CVPR,DanelljanBKF16} overcome this limitation by learning correlation filters on hierarchical deep features, but the real-time capacity largely fades away. Other deep trackers~\cite{WangLGY15,nam2016mdnet,Wang15} take full advantage of the end-to-end learning, and fine-tune the deep network online to achieve top performance. However, even on a high-end GPU, the speed of such trackers is usually around 1 fps which is too slow for practical use.

\begin{figure}[t]
\begin{center}
\includegraphics[width=0.9\linewidth]{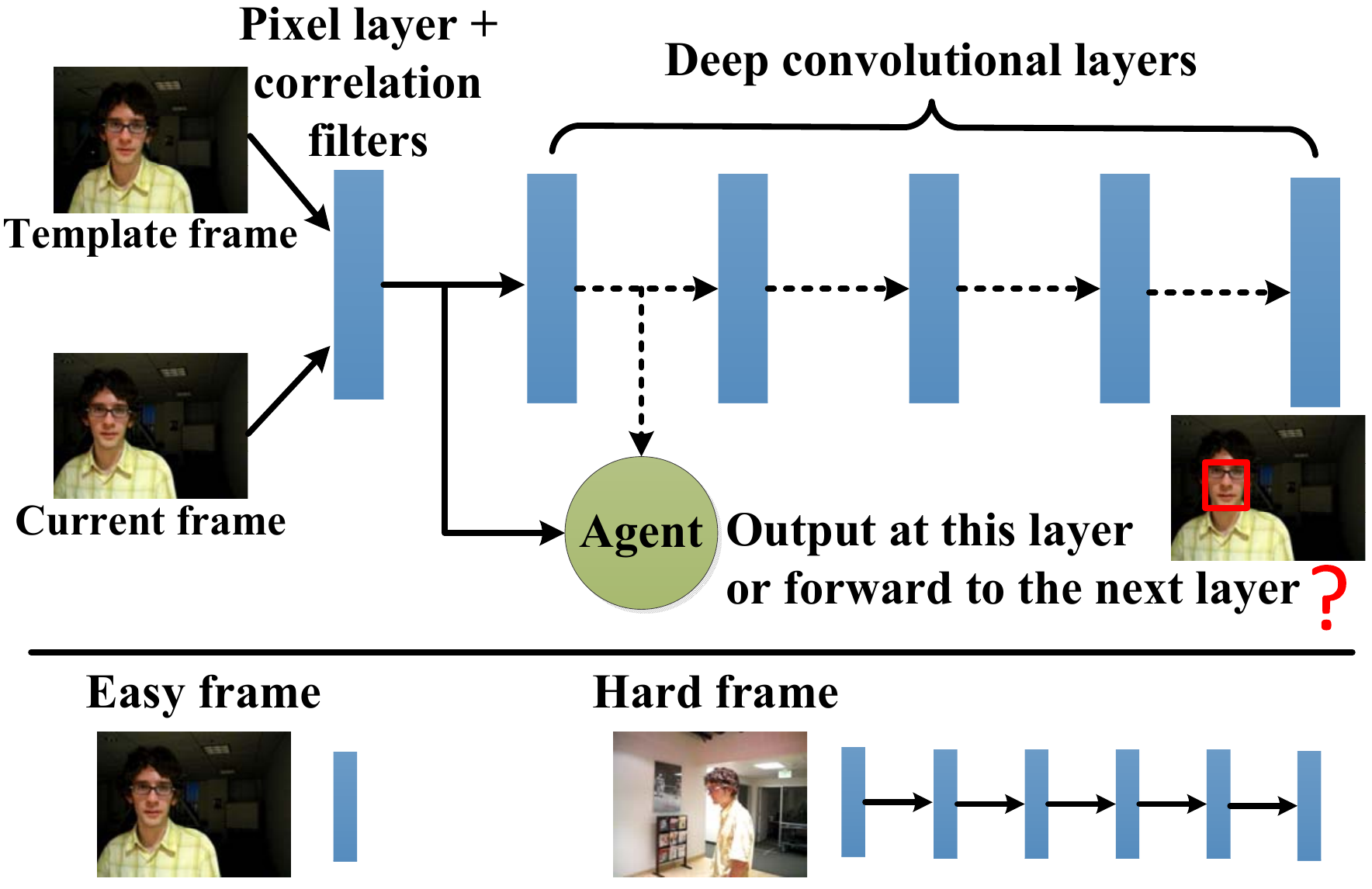}
\end{center}
\caption{Learning policies for adaptive tracking with deep feature cascades. The agent decides whether we can accurately locate objects on an early layer of the cascade. This avoids the need to always wait for the last layer to make decisions, saving a substantial amount of feed-forwarding time.}
\label{fig1}
\vspace{-1em}
\end{figure}

Two recent deep trackers~\cite{Bertinetto2016,held2016} significantly boost their speed by deploying fixed convolutional networks without any online learning. During tracking, the pre-trained network is simply traversed in a feed-forward pass for similarity evaluation or location regression, allowing to track objects at real-time speeds on GPU. Nevertheless, on a modern CPU, smartphone or tablet with much less computing power, such deep trackers can only process a couple of frames per second, far below the normal video frame-rate 30 fps. Obviously, the major computational burden comes from the forward pass through the entire network, and can be larger with deeper architectures.

We aim to improve both the accuracy and speed of deep tracking, instead of trading off the quality for speed by~\eg using compressed models~\cite{8897859}. We propose to {\em learn} to speed up deep trackers in an adaptive manner. Our adaptive approach builds on the observation that the tracking complexity varies across frames. For example, it is usually effective to use features from the last layer of a deep network to track objects undergoing large appearance change (\eg~abrupt motion) - as these higher-level features are more tolerant to dramatic appearance variation~\cite{MaICCV2015}. However, when the object is visually distinct or barely moves, early layers are in most scenarios sufficient for precise localization - offering the potential for substantial computational savings. In the extreme case, the ``zeroth'' pixel-level layer may suffice for such easy frames, as evidenced by the success of pixel-based correlation filters.

Such an adaptive strategy crucially depends on making the right {\em decision} - should the tracker stop at the current feature layer or keep computing features on the next layer? To this end, we learn an agent to automatically achieve this goal as illustrated in Fig.~\ref{fig1}. The agent learns to find the target at each layer, and decides if it is confident enough to output and stop there. If not, it moves forward to the next layer to continue. This is equivalent to learning a ``timely'' decision policy to choose the optimal layer for tracking. We will show such policy learning is much more robust than heuristically thresholding the response map of the current active layer. It is also in sharp contrast to the layer selection mechanism in~\cite{Wang15}, which only selects from two fixed levels of convolutional layers after the entire forward pass is finished. Instead, we formulate this problem as a decision-making process, making {\em sequential} decisions with early stopping ability. Specifically, we learn the policy in a reinforcement learning~\cite{mnih2015} fashion during the training phase, and simply apply the policy for adaptive tracking at test time.

In doing so, we are able to provide a speedup of around 10$\times$ (on average) over the baseline deep tracker~\cite{Bertinetto2016} and achieve even higher accuracies on existing OTB~\cite{7001050} and VOT~\cite{Kristan2015a} tracking benchmarks. Perhaps most tellingly, our approach provides a 100$\times$ speedup almost 50\% of the time, indicating the power of an adaptive approach - it turns out that most frames are quite easy to track! Accuracy is improved because each layer of the network is directly trained to be informative for adaptive tracking, similar to the past approaches for ``deep supervision''~\cite{LeeXGZT15}. Concretely, our adaptive tracker works by defining object templates across multiple layers of a network, including the ``zeroth'' pixel layer. Templates are evaluated across Region of Interest (ROI) with a convolutional filter. We use fast correlation filters~\cite{henriques2015} to compute response maps for lower layers of our network, where Fourier processing significantly speeds up the convolutional procedure (Fig.~\ref{fig1}). We refer to our approach as EArly-Stopping Tracker (EAST). On a single CPU, it has a near real-time average speed of 23.2 fps, and is about 190 fps almost 50\% of the time. This makes it the first CPU-friendly deep tracker among the top benchmark performers. It is worth noting that our policy learning method is quite generic. Further, it is readily applied to train end-to-end with an existing deep network for other time-critical vision tasks besides visual tracking.

%------------------------------------------------------------------------
\section{Related Work}
\noindent
{\bf Real-time tracking and correlation filters:} Visual tracking methods can rely on either generative (\eg~\cite{Ross2008}) or discriminative (\eg~\cite{6126251}) models. Discriminative models are often found to outperform in accuracy by discriminating the target from background. Such trackers can usually run fast using hand-crafted features (\eg~HOG~\cite{Dalal2005}) and various learning methods of P-N learning~\cite{6104061}, structured SVM~\cite{6126251}, multi-expert entropy minimization~\cite{zhang2014meem}, and correlation filter~\cite{Bolme10}.

Among them, Discriminative Correlation Filter (DCF)-based methods~\cite{Bolme10,henriques2015} are the family of tracking methods with high efficiency and high accuracy as well. The fast speed of DCF is due to its efficient use of all spatial shifts of a training sample by exploiting the discrete Fourier transform. The pioneering MOSSE~\cite{Bolme10} and improved Kernelized Correlation Filter (KCF)~\cite{henriques2015} trackers can operate at 669 fps and 292 fps respectively on a single CPU, which far exceeds the real-time requirement. Recent advances of DCF have achieved great success by the use of multi-feature channels~\cite{Danelljan14,Danelljan2016,MaICCV2015,Danelljan15,Qi_2016_CVPR,DanelljanBKF16}, scale estimation~\cite{DanelljanBMVC,Li2015}, long-term memory~\cite{MaYZY15}, and boundary effect alleviation~\cite{Danelljan15,Danelljan857265}. However, with increasing accuracy comes a dramatic decrease in speed (0.3--11 fps on a high-end GPU).

% sophicificated learning methods: Adaptive decontamination of the training set: A unified formulation for discriminative visual tracking.

\noindent
{\bf Tracking by deep learning:} Directly applying correlation filters on the multi-dimensional feature maps of deep Convolutional Neural Networks (CNNs) is one straight-forward way of integrating deep learning for tracking. Usually the deep CNN is fixed, and the DCF trackers trained on every convolutional layer are combined by a hierarchical ensemble method~\cite{MaICCV2015} or an adaptive Hedge algorithm~\cite{Qi_2016_CVPR}. Danelljan~\etal~\cite{Danelljan2016} recently introduced a continuous spatial domain formulation C-COT, to enable efficient integration of multi-resolution deep features. C-COT and its improved ECO~\cite{DanelljanBKF16} can achieve top performance in the VOT challenge~\cite{Kristan2015a}, but the tracking speed is still slow due to the high dimension of the entire deep feature space.

Another category of deep trackers~\cite{WangLGY15,nam2016mdnet,Wang15} update a pre-trained CNN online to account for the target-specific appearance at test time. Such trackers usually take a classification approach to classify many patches and choose the highest scoring one as the target object. Unfortunately, online training and exhaustive search severely hamper their speed. The top performing tracker MDNet~\cite{nam2016mdnet} has the GPU speed of only 1 fps or so. Recent advances include using Recurrent Neural Networks (RNNs)~\cite{KahouMM15,Cui_2016_CVPR} to model temporal information using an attention mechanism, but the speed remains slow.

One common reason for the slow speed of the above-mentioned deep trackers is they always conduct a complete feed-forward pass to the last CNN layer. This ignores the fact that the tracking complexity differs across varying conditions. One of our conclusions is that most frames in current video benchmarks are rather easy. For those frames, forwarding to only early layers may suffice. In principle, such an insight can be used to speed up many recent real-time deep trackers, such as GOTURN~\cite{held2016} (165 fps on GPU) and SiamFC~\cite{Bertinetto2016} (86 fps on GPU), to make them more CPU-friendly with near frame-rate speed.

\noindent
{\bf Feature selection in tracking:} Good features are important to tracking. The initial DCF trackers were limited to a single feature channel,~\eg~a grayscale image in MOSSE~\cite{Bolme10}. The DCF framework was later extended to multi-channel features such as HOG~\cite{DanelljanBMVC,henriques2015}, Haar-like features~\cite{6126251}, binary patterns~\cite{6104061} and Color Attributes~\cite{Danelljan14}. Generally, the hand-crafted features are cheap to compute, but they are not discriminative enough to handle complex tracking scenarios. Many recent deep trackers (\eg~\cite{WangLGY15,nam2016mdnet}) exploit the semantically robust features from the last CNN layer (fully-connected). However, spatial details of the tracked object are lost in the last layer which is not optimal for visual tracking. Danelljan~\etal~\cite{Danelljan15} found the first convolutional layer is very suitable for tracking. Other works~\cite{Danelljan2016,MaICCV2015,Qi_2016_CVPR,DanelljanBKF16} choose to utilize all the hierarchical convolutional features, where early layers can preserve high spatial resolution and deep layers are more discriminative.

In this paper, we make the best of hand-crafted and deep convolutional features in a cascaded structure, and learn an agent to select a minimum sequence of feature layers for fast tracking purposes. Unlike FCNT~\cite{Wang15} that selects features from two pre-defined layers only after a complete forward pass, our selection is sequential and can stop early at any layer with enough confidence.

\noindent
{\bf Feature cascades:} CNNs are a natural cascaded architecture with increasingly abstract feature representations. Contemporary works either improve the cascade's optimality by deep layer supervision~\cite{LeeXGZT15}, or stack multiple CNNs into a deeper cascade for coarse-to-fine~\cite{6619290} or multi-task~\cite{7780712} predictions. Our work differs in learning a decision policy of using only early feature layers in a cascade, and in combining feature cascades and reinforcement learning~\cite{sutton2012reinforcement} to achieve this goal. Our approach bears some similarity to the ``attentional cascade'' structure~\cite{viola2001rapid} which uses a cascade of gradually more complex classifiers. The difference is that attentional cascade aims to use early classifiers to eliminate easy negatives and reduce the burden of complex classifier evaluation, whereas we aim to use these early layers to make a strong decision as early as possible.

\noindent
{\bf Reinforcement learning for tracking:} Reinforcement learning (RL)~\cite{mnih2015,sutton2012reinforcement} is capable of learning good policies to take a sequence of actions based on trail and error. It has been successfully applied to vision tasks (\eg~object detection~\cite{Caicedo2015}) by treating them as a decision-making process. For visual tracking, there are two recent works that use RL to temporally attend target regions~\cite{ZhangMWW17} and select suitable template~\cite{Choi2017}. Our work is the first one to use RL to learn an early decision policy for speeding up deep tracking.

%------------------------------------------------------------------------
\section{Method}

\begin{figure*}[t]
\begin{center}
\includegraphics[width=1.0\linewidth]{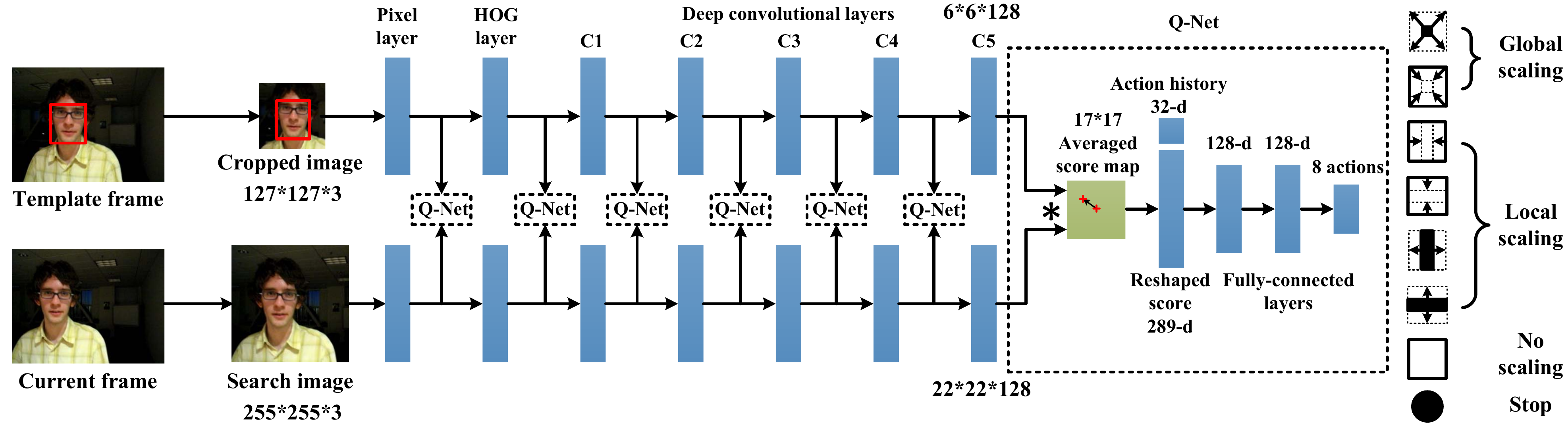}
\end{center}
\caption{System framework of our EArly-Stopping Tracker (EAST) by policy learning.}
\label{fig2}
\end{figure*}

\begin{figure*}[t]
\begin{center}
\includegraphics[width=1.0\linewidth]{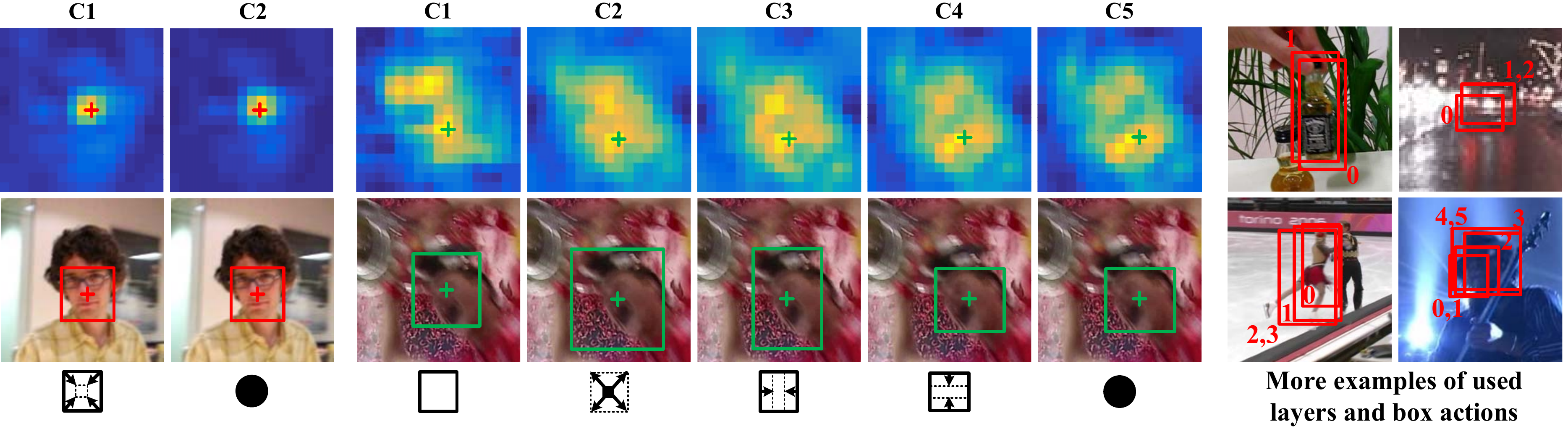}
\end{center}
\caption{Example sequences of actions taken to adjust the bounding box scale over deep convolutional layers. Box translation is determined by the relative position of maximum score on the score map. Note each score map is averaged with all the maps from preceding layers. Our agent learns to wisely act upon the score maps. It terminates the search early if the confidence score is highly peaked. When the score map is ambiguous (\eg~has two peaks for a blurry face in cluttered backgrounds), the agent postpones decision and enlarges the box according to the more unambiguous score map at the next layer. Further actions of box scaling are performed with stronger confidence.}
\label{fig3}
\vspace{-1em}
\end{figure*}

We use the deep Convolutional Neural Network (CNN) as a rich feature cascade for robust visual tracking. Given an input video frame at time $t$, the process of tracking an object with bounding box can be seen as the problem of maximizing a confidence function $f_l:H_t\rightarrow \mathbb{R}$ over the set of hypothesized object regions $R_t$:
\begin{equation}
\label{eq1}
r_t^{\ast} = \arg \max_{r\in H_t} f_l(r),
\end{equation}
where $H_t$ usually consists of regions around the object location in the previous frame, and $l$ denotes the feature layer at which the object confidence is obtained. Many deep trackers~\cite{WangLGY15,nam2016mdnet} exploit $l=L$ as the last fully-connected layer or last convolutional layer of AlexNet~\cite{NIPS2012_4824} or deeper VGG-Net~\cite{Simonyan14c}. Other trackers~\cite{Danelljan2016,MaICCV2015,Qi_2016_CVPR,DanelljanBKF16} exploit a full layer set $\{l\}_{l=1}^L$ of all the convolutional layers to take the best advantage of feature hierarchies. While these methods have been successful and effective, they are still slow and may not be needed when tracking during easy frames. % for proceeding through every layer to decide confidence, independent of the complexity of the object being tracked.

In this paper, we propose a principled sequential method that accumulates confidence $f_{l=1,\dots,L_t}(r)$ from an adaptively small set of feature layers $\{l\}_{l=1}^{L_t}$ in order to track efficiently without losing accuracy. For example, $L_t=2$ convolutional layers $C1$--$C2$ can suffice to track a distinct face in Fig.~\ref{fig3}; but for a blurry face in cluttered backgrounds, we may want to gather more evidence from a deeper layer $C5$. During such sequential search, our method needs to adjust the bounding box to progressively localize the object using more and more robust features. Ideally, we would like to minimize the number of forwarded layers necessary to locate objects. A naive approach might use a heuristic for determining when to advance to the next layer: for example, one might advance if the maximum value of the current response map is below a threshold. However, defining such heuristics can be difficult when the response map is ambiguous or has multiple peaks (\eg~Fig.~\ref{fig3}). Instead, we propose to train a functioning agent end to end by deep reinforcement learning~\cite{mnih2015}. The agent learns the action and search policy (including the early stopping criteria) so that it can make decisions under uncertainty to reach the target. Fig.~\ref{fig2} shows the framework of our policy learning.

\subsection{Fully-Convolutional Siamese Network}

In this section, we review the fully-convolutional Siamese tracker~\cite{Bertinetto2016}, which we make use of for its good trade-off between speed (86 fps on GPU) and accuracy in exploiting deep convolutional layers. Other trackers are either too slow~\cite{Danelljan2016,MaICCV2015,Qi_2016_CVPR,DanelljanBKF16} as a baseline ($<$11 fps on GPU), or do not provide explicit response maps (direct regression instead in~\cite{held2016}) for our policy learning purposes.

The Siamese network~\cite{Bertinetto2016} is trained offline to locate a $127\times 127$ template image $z$ within a larger $255\times 255$ search image $x$. A similarity function is learned to compare the template image $z$ to a candidate region of the same size in the search image $x$, so as to return a high score for the truth region and low score otherwise. Such similarity evaluation is fully-convolutional with respect to $x$ in the network, much more efficient than exhaustive search. Specifically, a cross-correlation layer is proposed to compute similarity for all translated sub-regions in $x$ in one pass:
\begin{equation}
\label{eq2}
F_l(z,x) = \varphi_l(z) \ast \varphi_l(x)+v\mathbb{I},
\end{equation}
where $\varphi_l$ is the convolutional feature embedding at layer $l=5$ (\ie~$C5$ layer), and $v\in \mathbb{R}$ is an offset value. Here $F_l(\cdot,\cdot)$ is a confidence score map of size $17\times 17$, as opposed to the single confidence score $f_l(\cdot)$ in Eq.~\ref{eq1}.

During tracking, this Siamese network simply evaluates the similarity online between the template image in previous frame and the search region in current frame, leading to fast speed. The relative position of the maximum score multiplied by the stride of the network, gives the object translation from frame to frame.

Our goal is to learn an early decision policy from these confidence score maps $F_l$, to adaptively 1) predict the object bounding box across layers, and 2) stop early at a layer $l<5$ when sufficiently confident about the prediction. Note the score map dimension depends on the size of input feature maps. The score map $F_l$ on an early layer $l$ will have a larger resolution than $17\times 17$, and so we downsample to this size to facilitate learning. Also, SiamFC~\cite{Bertinetto2016} searches over multiple scales of the search image to handle scale variations. We only work on the {\em original} scale for high efficiency, and learn to gradually infer the box scale from heatmaps computed during the single forward-pass.

\subsection{Learning Policies with Reinforcement Learning}

We treat the tracking problem as a Markov Decision Process (MDP) where an agent can make a sequence of actions across feature layers, see Fig.~\ref{fig2}. This agent learns when to stop advancing to the next layer, as well as how to to gradually deform a bounding box once per layer, which is initialized to the estimated box from the previous frame. The ultimate goal is to output a tight box around the object with as few as layers as possible. The challenge is to be able to operate with rewards that rule out supervision at each step, and at the same time, to minimize the number of steps to locate the target given its changing complexity.

We train the agent in a reinforcement learning (RL) setting to learn decision policies. In the RL setting there are a set of states $S$ and actions $A$, and a reward function $R$. At each step on layer $l$, the agent examines the current state $S_l$, and decides on the action $A_l$ to either transform the box or {\em stop} with a box output. The action $A_l$ is expected to reduce the uncertainty in localizing the object, and receives positive or negative reward $R_l$ reflecting how well the current box covers the object and how few steps are used before action {\em stop}. By maximizing the expected rewards, the agent learns the best policy to take actions and can explicitly balance accuracy (search for more layers) and efficiency (stop early if highly confident).

\noindent
{\bf Actions:} Our action set $A$ includes seven anisotropic scaling transformations to a box and one {\em stop} action to terminate search. We do not use the agent to predict the centroid of the box, and instead compute it directly from the relative position of the maximum score on score map as in Eq.~\ref{eq1}. We also experimented with requiring the agent to report box translations, but found directly inferring them from the score maps simplified training and increased convergence (due to the smaller space of actions $A$). %In our experiments, we found this deterministic way can already well estimate the translation of objects with often small displacements between frame. Besides, excluding box translation from $A$ largely reduces the action space to be learned by an agent, making its training much easier.

For scaling actions, there are two global and four local (modify aspect ratio) transformations as shown in Fig.~\ref{fig2}. Similar to~\cite{Caicedo2015}, any of these actions makes a combined horizontal and vertical change to the box by a factor of 0.2 relative to its current size. We also introduce a {\em no scaling} action that does not scale the box at all. This action allows the agent to postpone decision when the current score map is ambiguous or a decision simply cannot be made. Fig.~\ref{fig3} exemplifies this case where two peaks exist on the first-layer score map for a cluttered scene. The agent decides not to act on this map but waits for a more unambiguous map at the next layer to act (enlarge the box).

\noindent
{\bf States:} The state $S_l$ is represented as a tuple $(F'_l,h_l)$, where $F'_l$ is the score map and $h_l$ is a vector of the history of taken actions. We define $F'_l=\sum_{k=1}^l F_k /l$ as the average of the score map at current layer $l$ and all its preceding maps from earlier layers. Thus $F'_l$ not only encodes the currently observed confidence but also the confidence history that has been collected. This is found to work better empirically than using $F_l$ only, and is similar to the hypercolumn representation~\cite{7298642} with the benefits of simultaneously leveraging information from early layers that capture fine-grained spatial details and deeper layers that capture semantics. Also, the resulting robustness comes at a negligible cost in averaging the score maps already obtained. The history vector $h_l$ keeps track of the past 4 actions. Each action in the vector is represented by an 8-dimensional one-hot vector or zero vector (when processing the first layer). We find that including  $h_l\in \mathbb{R}^{32}$ helps stabilize action trajectories.

\noindent
{\bf Rewards:} The reward function $R(S_{l-1},S_{l})$ reflects the localization accuracy improvement from state $S_{l-1}$ to $S_{l}$ after taking a particular action $A_l$. The accuracy is measured by the Intersection-over-Union (IoU) between the predicted box $b$ and ground-truth box $g$. We can formally define IoU as $IoU(b,g)=area(b\cap g)/area(b\cup g)$. Since each state $S_l$ is associated with a box $b_l$, the reward function is then defined following~\cite{Caicedo2015}:
\begin{equation}
\label{eq3}
R(S_{l-1},S_{l})\!=\!
\begin{cases}
\mathrm{sign}(IoU(b_{l},g)-IoU(b_{l-1},g)),& \!\!\!\!\! A_l \ne stop\\
+3, \qquad \qquad \quad \!\!\!\!\!\!\!\! IoU(b_{l},g)\ge 0.6,& \!\!\!\!\! A_l = stop\\
-3, \qquad \qquad \quad \!\!\!\!\!\!\!\! IoU(b_{l},g)< 0.6,&  \!\!\!\!\! A_l = stop,\\
\end{cases}
\end{equation}
where the accuracy improvement is quantized to $\pm1$ if the current action is not {\em stop}. This reward scheme encourages positive transformations even with small accuracy improvement. If there is no transformation for further improvement, or if the agent already arrives at the last layer $l=L$, the action should be {\em stop}. In this case, the reward function will penalize the predicted box $b_l$ with IoU less than 0.6. Note such a reward scheme implicitly penalizes a large number of layers $l$ since Q-learning (detailed next) models the expected future rewards when deciding on an action (positive or negative).

\noindent
{\bf Deep Q-learning:} The optimal policy of selecting actions should maximize the sum of expected rewards on a given frame. Since we do not have {\em a priori} knowledge about the correct layer or action to choose, we address the learning problem through  deep Q-learning~\cite{mnih2015}. This approach learns an action-value function $Q(S_l,A_l)$ to select the action $A_{l+1}$ that gives the highest reward at each layer. The learning process iteratively updates the action-selection policy by:
\begin{equation}
\label{eq4}
Q(S_l,A_l)=R+\gamma \max_{A'}Q(S',A'),
\end{equation}
where $Q(S',A')$ is the future reward and $\gamma$ the discount factor. The function $Q(S,A)$ is learned by a deep Q-Network as illustrated in Fig.~\ref{fig2}. It takes as input the state representation $S$,~\ie~the reshaped score vector and action history vector. The network consists of two 128-dimensional fully-connected layers, finally mapping to 8 actions. Each fully-connected layer is randomly initialized, and is followed by ReLU and dropout regularization~\cite{NIPS2012_4824}.

Note during training, we not only update the weights of the Q-Network, but also the pre-trained convolutional layer when the agent receives rewards on that layer. Similar to the deeply-supervised net~\cite{LeeXGZT15}, our approach provides a direct target signal for learning the feature representation at each layer, so as to improve performance of our adaptive tracker. %hidden layer supervision on side responses makes each layer more reliable for tracking and improves performance.

\noindent
{\bf Testing with learned policies:} During testing, the agent does not receive rewards or update the Q-function. It just follows the decision policy to deform the box and output it when a {\em stop} action is performed. Our agent takes only 2.1 steps on average to locate the target between frames on the OTB-50 dataset~\cite{WuLimYang13}. This means we can correctly track most objects by using 2 deep layers. Only for those hard frames, the search degenerates to a full forward pass. The overall tracking algorithm follows SiamFC~\cite{Bertinetto2016} to search over candidate regions around the estimated location from previous frame. As a result, one order of magnitude speed-ups are achieved over traditional non-adaptive deep trackers. Set aside efficiency, the policies are still appealing in that they mimic the dynamic {\em attention} mechanism by progressively attending to the target region in feature cascades.
%This significantly differs from regression methods~\cite{held2016} that produce a single structured prediction.

\noindent
{\bf Implementation details:} We use the AlexNet~\cite{NIPS2012_4824}-like convolutional architecture as in SiamFC~\cite{Bertinetto2016}. The whole network including Q-Net is trained on the ImageNet Video~\cite{ILSVRC15} trainval set (4417 videos) for 50 epochs, each completed after the agent has interacted with all training images.  We make use of an $\epsilon$-greedy~\cite{sutton2012reinforcement} optimization during Q-learning, taking a random action with probability $\epsilon$ to encourage exploration of diverse action policies. We anneal $\epsilon$ linearly from 1 to 0.1 over the first 30 epochs, and fix $\epsilon$ to 0.1 in the remaining 20 epochs. We use a learning rate $1e$-3, discount factor $\gamma=0.9$ and batch size 64. The network parameters are updated with direct stochastic gradient descent using MatConvNet~\cite{Vedaldi15} on a single NVIDIA GeForce Titan X GPU and an Intel Core i7 CPU at 4.0GHz.

\subsection{Learning with cheap features}
Our policy learning can be applied to a feature cascade with any type of feature layers. We explore the use of additional cheap feature layers after the pixel-layer and before the more expensive deep layers. Inspired by the success of correlation filters defined on multi-channel HOG layers~\cite{Dalal2005}, we explore an optional HOG layer. In theory, other cheap features such as Color Attributes~\cite{Danelljan14} might apply. When processing our pixel and HOG layers, we make use of fast correlation filters. We specifically make use of the Dual Correlation Filter (DCF)~\cite{henriques2015}, which exhibits a good tradeoff in CPU speed (270+ fps on CPU) and accuracy compared to alternatives such as the Kernelized Correlation Filter (2$\times$ slower), STC~\cite{Zhang2014} (350 fps but lower accuracy) and SRDCF~\cite{Danelljan857265} (5 fps).

\section{Results}

\begin{figure*}[t]
\begin{center}
\includegraphics[width=1.02\linewidth]{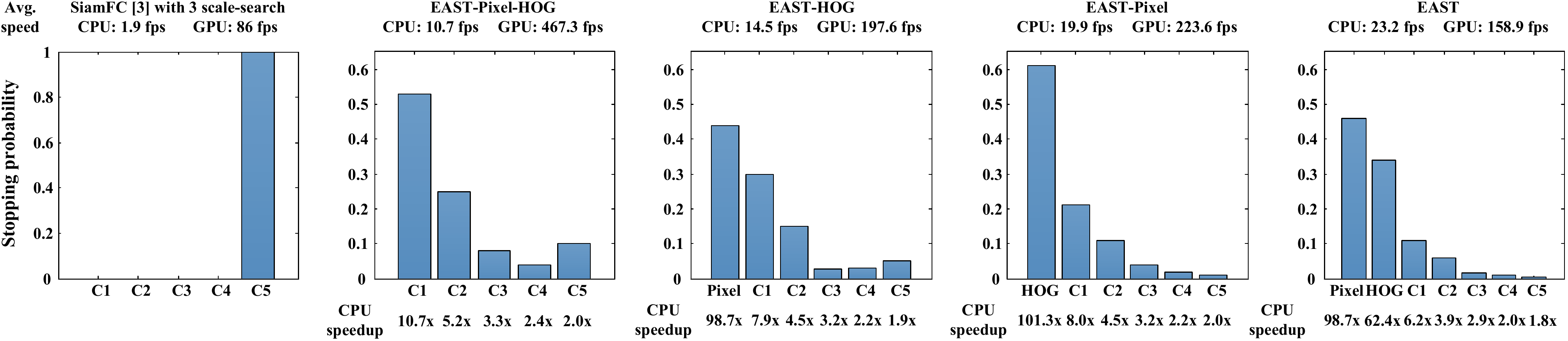}
\end{center}
\caption{Stopping probability at different feature layers and the resulting speed on OTB-50 dataset~\cite{WuLimYang13}. For each of the 5 models with different feature cascades, we show the average speed on both CPU and GPU (top), and the CPU speedup ratio over the baseline SiamFC~\cite{Bertinetto2016} (1.9 fps) at each layer (bottom). SiamFC searches over multi-scaled images to handle scale variations, while we predict the scale in a single forward-pass, leading to a constant tracking speedup. Our early stopping policy further accelerates tracking (4.5$\times$) by adaptively using early layers as compared to SiamFC that always uses the last layer C5. When the early layer is the cheap HOG or pixel layer with fast CPU speed (270+ fps), we are able to increase the average CPU speed by one order of magnitude and operate at around 100$\times$ faster speed for almost 50\% of the time. Our full model EAST operates at near real-time speed 23.2 fps on CPU. On the other hand, it is found the more reliance on CPU computations will generally increase the CPU speed, but also lose the benefits of GPU speed to some acceptable extent.}
\label{fig4}
\vspace{-1em}
\end{figure*}

\begin{figure}[t]
\begin{center}
\includegraphics[width=1.03\linewidth]{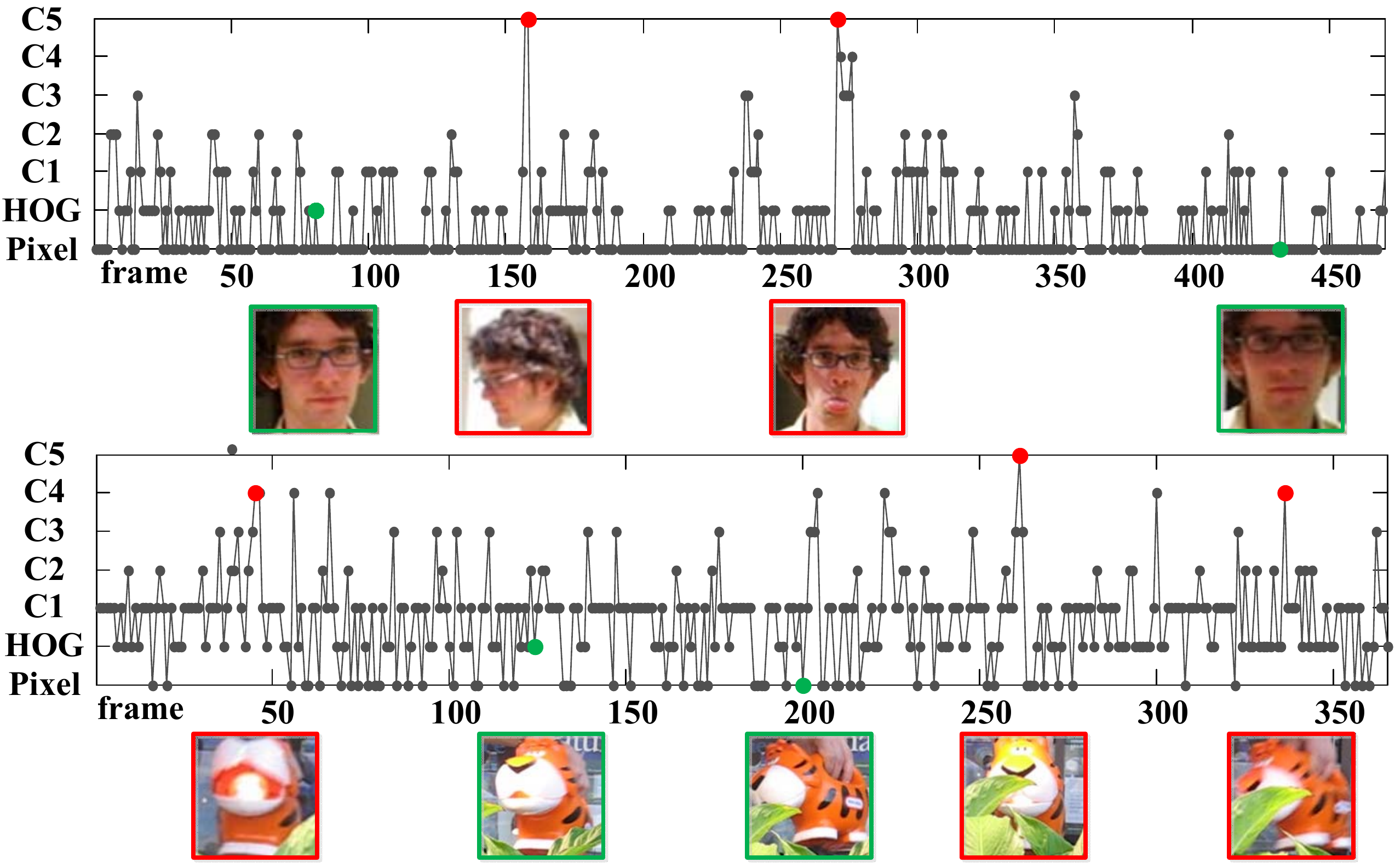}
\end{center}
\caption{Stopped feature layers for the frames in videos {\em David} (frames 300-770 of the original video) and {\em Tiger2}. Easy and hard frames are marked in green and red.}
\label{fig5}
\vspace{-1em}
\end{figure}

Before comparing our EArly-Stopping Tracker (EAST) with prior works, we first conduct an ablation study of some of its important variants. We compare EAST to variants using different feature cascades---EAST-Pixel-HOG, EAST-HOG, EAST-Pixel, as well as the baseline SiamFC~\cite{Bertinetto2016}.  Fig.~\ref{fig4} shows the probability of stopping at different feature layers and the associated speed on OTB-50~\cite{WuLimYang13}. EAST indeed learns to use only early layers over 70\% of the time, as they suffice when tracking easy frames. Hard frames are processed with additional layers, degenerating to a full forward-pass (as in SiamFC) only when needed. Fig.~\ref{fig5} illustrates some examples of easy and hard frames (and their stopped layers) on particular video sequences.

Obviously, the earlier layer we stop at, the greater the speedup. EAST-Pixel-HOG  (which lacks a pixel or HOG layer) is about 4.5$\times$ faster than SiamFC on average, running at 10.7 fps on a CPU and 467.3 fps on a GPU. Recall that Pixel and HOG layers can be processed by fast correlation filters~\cite{henriques2015} that run at 278 fps and 292 fps, respectively. By adding such layers, EAST achieves a CPU speedup of $10\times$ on average and $100\times$ for those easy frames. Even though HOG may incur an additional feature computation cost compared to the Pixel layer, it produces a greater speedup because it enables more accurate pruning, and so is selected more often by EAST. Note that the average GPU speed is sacrificed to some extent due to larger reliance on CPU computations. Nonetheless, EAST still produces a near-real-time CPU rate of 23.2 fps, making it quite practical for CPU-bound tracking (required on many embedded devices).

Table~\ref{tb1} (top) summarizes the speed and accuracy of these EAST variants. We use the Overlap Success (OS) rate as a strict evaluation metric for accuracy. One-Pass Evaluation (OPE) is employed to compare accuracies in terms of Area Under the Curve (AUC) of OS rates for different threshold values. Table~\ref{tb1} shows that the use of {\em more} and {\em stronger} feature layers systematically improves the accuracy of the tracker, reaching to the AUC score of 0.638 for our full EAST model. Speed is also improved due to the larger computational savings on cheap layers.

Table~\ref{tb1} (middle) further compares other EAST variants:
\vspace{-1.2em}
\begin{packed_itemize}
\item EAST$_{last}$: tracking by always forwarding to the last feature layer. 
\item EAST$_{th}$: feed-forwarding if the maximum value of the current response map is lower than 0.9. Scale is determined by the size of thresholded region.
\item EAST$_{-ch}$: policy learning without confidence history,~\ie~we use $F_l$ instead of $F'_l=\sum_{k=1}^l F_k /l$.
\item EAST$_{-ah}$: policy learning without action history $h_l$.
\end{packed_itemize}
\vspace{-0.3em}

EAST$_{last}$ is similar to the baseline SiamFC~\cite{Bertinetto2016} in that all layers are always processed, but differs in that it is trained with deep supervision and does not require an image pyramid to model scale. As a result, it both works better and faster (in CPU terms) than SiamFC. Interestingly, EAST$_{last}$ outperforms EAST but is much slower. This suggests that our adaptive strategy is slightly hurting accuracy. EAST$_{th}$ makes use of manually-designed heursitics for stopping, avoiding the need to predict actions with a Q-Net, making it 2$\times$ faster on CPU. However, heuristic policies fail to work as well as the one learned by Q-learning. Finally, eliminating scoremap and action histories also hurts performance, likely because such histories stabilize the search process. % and action history in methods EAST$_{-ch}$ and EAST$_{-ah}$, we observed that both accuracy and speed (due to longer unstable search) are hurt. 

% Lastly, we remark that we can use Fourier processing on those deep layers for which its faster than convolution (can be evaluated on a per-layer basis). Plus, the template can now be updated online.
% Mention that fourier processing may also speed up lower-layers from deep network

\begin{table}[t]
\caption{One-Pass Evaluation (OPE) results of Area Under the Curve (AUC) score and speed (CPU/GPU fps) on OTB-50 dataset.}
\centering
\resizebox{0.65\linewidth}{!}{
\begin{tabular}{!{\vrule width1pt} c !{\vrule width1pt} c|c !{\vrule width1pt}}
    \Xhline{1pt}
    Method & AUC & Speed \\    \Xhline{1pt}
    EAST-Pixel-HOG & 0.619 & 10.7 / 467.3 \\
    EAST-HOG & 0.628 & 14.5 / 197.6 \\
    EAST-Pixel & 0.631  & 19.9 / 223.6 \\ \Xhline{1pt}
    EAST$_{last}$ & 0.645 & 3.4 / 76.3 \\
    EAST$_{th}$ & 0.597 & 45.7 / 172.0 \\
    EAST$_{-ch}$ & 0.621 & 21.1 / 137.2 \\
    EAST$_{-ah}$ & 0.610 & 17.4 / 127.6 \\ \Xhline{1pt}
    SiamFC~\cite{Bertinetto2016} & 0.612 & 1.9 / 86 \\ 
    EAST & 0.638 & 23.2 / 158.9 \\ \Xhline{1pt}
    \end{tabular}
}
%\end{center}
\label{tb1}
%\vspace{-1em}
\end{table}

\noindent
{\bf OTB-50 results:} The OTB-50~\cite{WuLimYang13} benchmark contains 50 video sequences for evaluation. Table~\ref{tb2} compares the AUC scores of our EAST and the state-of-the-art trackers: TLD~\cite{6104061}, GOTURN~\cite{held2016}, Struck~\cite{6126251}, KCF~\cite{henriques2015}, DSST~\cite{DanelljanBMVC}, MEEM~\cite{zhang2014meem}, RTT~\cite{Cui_2016_CVPR}, FCNT~\cite{Wang15}, Staple~\cite{Bertinetto16}, HDT~\cite{Qi_2016_CVPR}, HCF~\cite{MaICCV2015}, LCT~\cite{MaYZY15}, SiamFC~\cite{Bertinetto2016} and SINT~\cite{TaoCVPR2016}. CPU/GPU speeds (fps) are also reported.

EAST achieves the highest AUC of any method. It does so while being significantly faster. For example, the runner-up SINT is 4 fps on GPU, while other GPU-based real-time deep trackers (GOTURN and SiamFC)  are significantly slower on a CPU (2-3 fps). When compared to fast correlation trackers~\eg~KCF defined on cheap features, EAST is significantly more accurate while still maintaining near-real-time speeds. This is in contrast to correlation filters defined on deep features (HCF and HDF), who are not real-time even with a GPU. %Our early decision policies help escape from this dilemma in deep feature cascades. Note 
The Staple tracker combines HOG and color features by a ridge regression, while FCNT tracks by selecting features from deep layers. EAST outperforms both in terms of accuracy. %by {\em learning} all deep layers and sequential layer selection policies. 
The Recurrently Target-attending Tracker (RTT) trains Recurrent Neural Networks (RNNs) to capture attentions as a regularization on correlation filter maps. However, it is noticeably slower and less accurate than EAST (3 fps on CPU, and an AUC of 0.588).

\begin{table*}[t]
\caption{The Area Under the Curve (AUC) score for One-Pass Evaluation (OPE), and speed (fps, * indicates GPU speed, otherwise CPU speed) on the OTB-50 dataset. The best results are shown in bold.}
\centering
\resizebox{1.0\linewidth}{!}{
\begin{tabular}{!{\vrule width1pt} c !{\vrule width1pt} c|c|c|c|c|c|c|c|c|c|c|c|c|c|c !{\vrule width1pt}}
    \Xhline{1pt}
    Method & TLD & GOTURN & Struct & KCF & DSST & MEEM & RTT & FCNT & Staple & HDT & HCF & LCT & SiamFC & SINT & EAST \\
     & \cite{6104061} & \cite{held2016} & \cite{6126251} & \cite{henriques2015} & \cite{DanelljanBMVC} & \cite{zhang2014meem} & \cite{Cui_2016_CVPR} & \cite{Wang15} & \cite{Bertinetto16} & \cite{Qi_2016_CVPR} & \cite{MaICCV2015} & \cite{MaYZY15} & \cite{Bertinetto2016} & \cite{TaoCVPR2016} & \\    \Xhline{1pt}
    AUC & 0.438 & 0.450 & 0.474 & 0.516 & 0.554 & 0.572 & 0.588 & 0.599 & 0.600 & 0.603 & 0.605 & 0.612 & 0.612 & 0.625 & \textbf{0.638} \\
    Speed & 22 & \textbf{165}* & 10 & \textbf{172} & 24 & 10 & 3 & 3* & 80 & 10* & 11* & 27 & 86* & 4* & 23/159* \\ \Xhline{1pt}
    \end{tabular}
}
%\end{center}
\label{tb2}
\vspace{-1em}
\end{table*}

\begin{table}[t]
\caption{The Area Under the Curve (AUC) score for One-Pass Evaluation (OPE), and speed (fps, * indicates GPU speed, otherwise CPU speed) on the OTB-100 dataset. The best results are shown in bold.}
\centering
\resizebox{1.0\linewidth}{!}{
\begin{tabular}{!{\vrule width1pt} c !{\vrule width1pt} c|c|c|c|c|c !{\vrule width1pt}}
    \Xhline{1pt}
    Method & RDT & SRDCF & MDNet & C-COT & ECO & EAST \\
     & \cite{Choi2017} & \cite{Danelljan857265} & \cite{nam2016mdnet} & \cite{Danelljan2016} & \cite{DanelljanBKF16} &  \\    \Xhline{1pt}
    AUC & 0.603 & 0.605 & 0.685 & 0.686 & \textbf{0.694} & 0.629  \\
    Speed & 43* & 5 & 1* & 0.3 & 6 & \textbf{23/159}* \\ \Xhline{1pt}
    \end{tabular}
}
%\end{center}
\label{tb3}
\vspace{-1em}
\end{table}

\noindent
{\bf OTB-100 results:} The OTB-100~\cite{7001050} dataset is the extension of OTB-50 and is more challenging. We test on the full 100 videos to compare with recent related trackers: RDT~\cite{Choi2017}, SRDCF~\cite{Danelljan857265}, MDNet~\cite{nam2016mdnet}, C-COT~\cite{Danelljan2016}, ECO~\cite{DanelljanBKF16}. Table~\ref{tb3} summarizes their AUC scores and CPU/GPU speeds. EAST is close to state-of-the-art in terms of accuracy and is the fastest among the top performers on OTB-100. The MDNet and correlation-filter-based advances SRDCF, C-COT and ECO all suffer from low speed, while EAST does not sacrifice the run-time performance for accuracy. RDT is a related Reinforcement Learning (RL)-based method that selects the best template to track a given frame. EAST (which also makes use of RL) proves more accurate while being 4$\times$ faster on a GPU.

\begin{figure}[t]
\begin{center}
\includegraphics[width=0.75\linewidth]{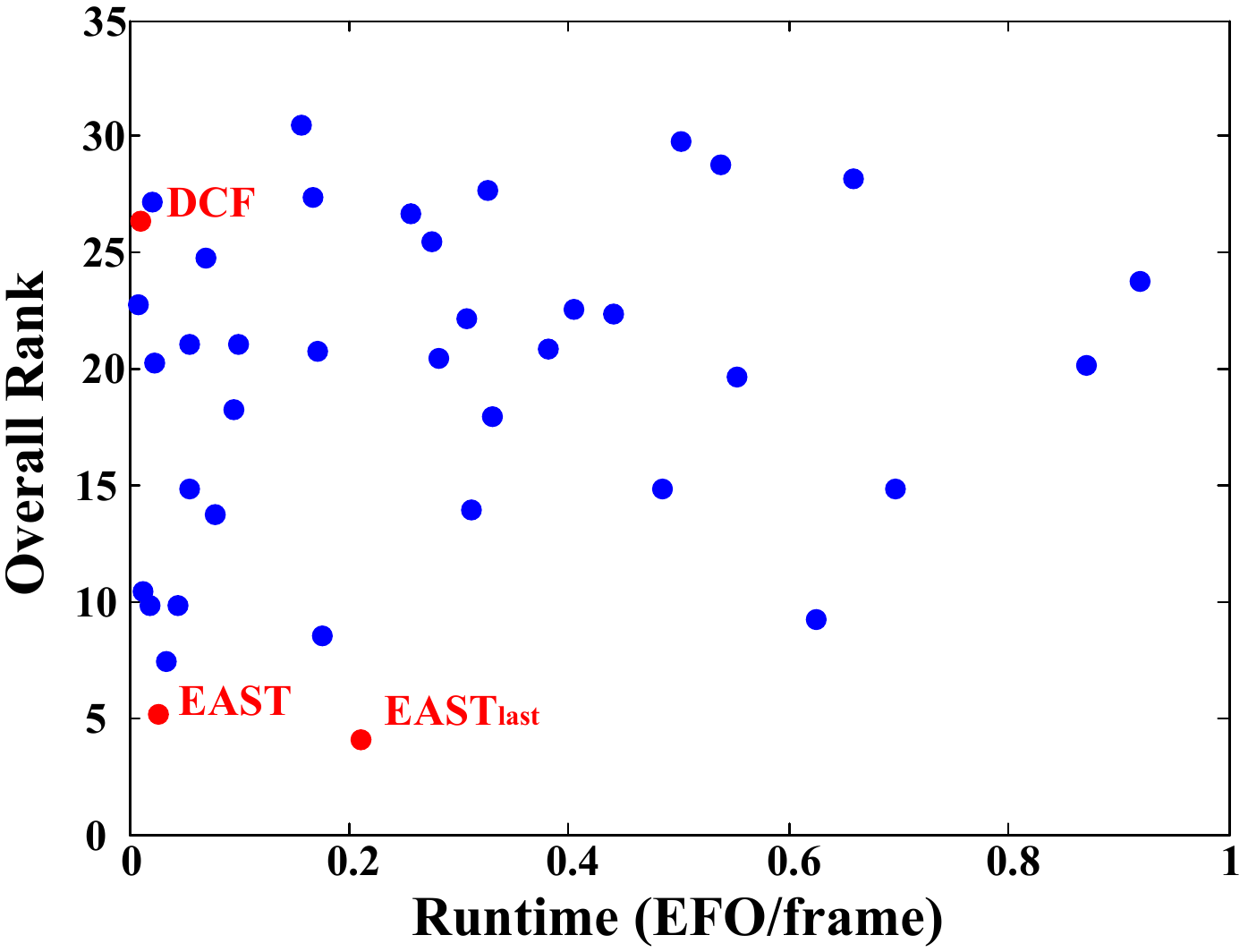}
\end{center}
\caption{Overall rank and runtime of our tracker (red) in comparison to the 38 trackers from VOT-14 Challenge. We show the results of EAST and its two variants that always forward to the first (\ie~DCF on pixel layer) and last (\ie~EAST$_{last}$) feature layer.}
\label{fig6}
\vspace{-1em}
\end{figure}

\noindent
{\bf VOT-14 results:} We test EAST on the 25 videos from the VOT-14~\cite{Hadfield14d} Tracking Challenge. Fig.~\ref{fig6} compares the speed and overall rank of EAST to that of the 38 tracker submissions. For the rank computation, the challenge first evaluates the average accuracy and robustness ranking (refer to the supplementary material for details) for each tracker, and then averages the two rank values to get the overall rank. As can be seen from Fig.~\ref{fig6}, EAST achieves the best accuracy-speed trade-off, outperforming all competitors in the overall rank. We also show two EAST variants at the opposite extreme: always forwarding to the first (\ie~DCF~\cite{henriques2015} on pixel layer) and last (\ie~EAST$_{last}$) feature layer. Our EAST is able to adaptively take advantage of the speed and accuracy benefits of two variants by policy learning.

\begin{table}[t]
\caption{Raw scores and speed for our method and the top 4 trackers of the VOT-15 Challenge. The * indicates the speed in EFO units. The CPU/GPU speeds for our EAST are given.}
\centering
\resizebox{0.95\linewidth}{!}{
\begin{tabular}{!{\vrule width1pt} c !{\vrule width1pt} c|c|c|c !{\vrule width1pt}}
    \Xhline{1pt}
    Tracker & Accuracy & Robustness & Overlap & Speed (fps) \\    \Xhline{1pt}
    MDNet~\cite{nam2016mdnet} & 0.60 & 0.69 & 0.38 & 1 \\
    EAST & 0.57 & 1.03 & 0.34 & \textbf{21/148} \\
    DeepSRDCF~\cite{Danelljan15} & 0.56 & 1.05 & 0.32 & $<$1* \\
	EBT~\cite{zhu2015tracking} & 0.47 & 1.02 & 0.31 & 5 \\
    SRDCF~\cite{Danelljan857265} & 0.56 & 1.24 & 0.29 & 5 \\\Xhline{1pt}
    \end{tabular}
}
%\end{center}
\label{tb4}
\vspace{-1em}
\end{table}

\noindent
{\bf VOT-15 results:} The VOT-15~\cite{Kristan2015a} Tracking Challenge has 60 testing videos chosen from a pool of 356. Trackers are automatically re-initialized five frames after failure (zero overlap). Table~\ref{tb4} compares our EAST with the top 4 trackers in terms of accuracy and speed (using the vot2015-challenge toolkit). Our testing speed on this benchmark are 21 fps on CPU and 148 fps on GPU, making EAST the fastest and most CPU-friendly tracker among the top performers. We achieve comparable accuracy scores to MDNet, while providing a 148$\times$ speedup on GPU, indicating the power of our adaptive policy learning approach.

\section{Conclusion}
This paper proposes an adaptive approach to tracking with deep feature cascades. Our fundamental insight is that most frames in typical tracking scenarios turn out to be easy, in that simple features (such as pixels or HOG) suffice. That said, some challenging frames do require ``heavy-duty" invariant feature processing. The challenge is in determining which is which! By formulating the tracking problem as a decision making process, we learn a reinforcement-learning agent that can make such distinctions. Importantly, the agent learns to do so in an iterative manner, making efficient use of a feature cascade that proceeds to deeper layers only when the current one does not suffice. This dramatically reduces the feed-forwarding cost for those easy frames (by 100X), leading to an overall significant speedup for tracking. Such a policy learning method is appealing in that it is trained end-to-end and can be applied to any deep network designed for time-critical tasks.

\noindent
{\bf Acknowledgment:} This research was supported in part by NSF grants CNS-1518865, IIS-1618903 and IIS-1526033 and the DARPA grant HR001117C0051. Additional support was provided by the Intel Science and Technology Center for Visual Cloud Systems (ISTC-VCS), Google, and Autel. Any opinions, findings, conclusions or recommendations expressed in this material are those of the authors and do not necessarily reflect the view(s) of their employers or the above-mentioned funding sources.

{\small
\bibliographystyle{ieee}
\bibliography{mybib}
}

\clearpage

\section*{Supplementary Material}

%%%%%%%%% BODY TEXT
\section*{A. Algorithmic Details}

\noindent
{\bf Network architecture:} We use the exact convolutional architecture of SiamFC~\cite{Bertinetto2016}. The convolutional layers $C1-C5$ and their parameter details are given in Table~\ref{tb5}. Note max-pooling is employed for the convolutional layers $C1$ and $C2$. We use the nonlinear ReLU function~\cite{NIPS2012_4824} after every convolutional layer except for $C5$. Batch normalization is inserted after every linear layer.

\noindent
{\bf Deep Q-learning:} During deep Q-learning~\cite{mnih2015}, the optimal action-value function $Q(S_l,A_l)$ obeys the Bellman equation: it is optimal to select the action $A'$ that maximizes the expected reward
\begin{equation}
\label{eq5}
Q(S_l,A_l)=R+\gamma \max_{A'}Q(S',A'),
\end{equation}
where $Q(S',A')$ is the future reward and $\gamma$ the discount factor. Since the action-value function is approximated by a Q-Net with weights $\theta$, the Q-Net can be trained by minimizing the loss function $V(\theta_l)$ at each iteration $l$,
\begin{equation}
\label{eq6}
V(\theta_l) = \mathbb{E} \left[ \left( R+\gamma \max_{A'}Q(S',A';\theta_{l-1})-Q(S_l,A_l;\theta_l) \right)^2 \right].
\end{equation}

The gradient of this loss function with respect to the network weights $\theta_l$ is as follows:
\begin{eqnarray}
\label{eq7}
\nabla_{\theta_l} V(\theta_l) = && \!\!\!\!\!\!\!\!\!\! \mathbb{E} \left[ \left( R+\gamma \max_{A'}Q(S',A';\theta_{l-1})-Q(S_l,A_l;\theta_l) \right) \right. \nonumber  \\ 
  &&  \cdot \nabla_{\theta_l} Q(S_l,A_l;\theta_l) \Big]. 
\end{eqnarray}

\section*{B. Discussions and Results}

The main idea of our EArly-Stopping Tracker (EAST) is to track easy frames using only early layers of a deep feature cascade,~\eg~pixel values, while hard frames are processed with invariant but expensive deep layers when needed.

An attached video {\bf demo.mp4} exemplifies such tracking policies in video sequences. To further validate the advantages of EAST in both accuracy and speed, we compare with the top 3 trackers on OTB-50~\cite{WuLimYang13} in terms of speed, and AUC score for One-Pass Evaluation (OPE), Temporal Robustness Evaluation (TRE) and Spatial Robustness Evaluation (SRE). Table~\ref{tb6} shows that EAST achieves the highest scores under all evaluation metrics, while maintaining fast tracking speed.

Table~\ref{tb7} shows the detailed ranks of accuracy $R_A$ and robustness $R_R$ under baseline and region noise experiments in VOT-14 challenge. The two experiments evaluate trackers with the initial target location from ground truth and that perturbed with random noises. The table also lists the overall rank $R_o$ and running speed to compare EAST with the best 3 trackers out of 38 submitted ones. It is evident that EAST is one of the fastest trackers, while outperforming other top performers in the overall rank.

\begin{table}[t]
\caption{Network architecture and convolutional layer specifics.}
\centering
\resizebox{1.0\linewidth}{!}{
\begin{tabular}{!{\vrule width1pt} c c|c|c|c|c|c !{\vrule width1pt}}
    \Xhline{1pt}
    \multicolumn{2}{!{\vrule width1pt}c|}{\multirow{2}{*}{Layer}} & \multirow{2}{*}{Support} & \multirow{2}{*}{Stride} & Template & Search & \multirow{2}{*}{Chans.} \\ 
    &  &  &  & activation & activation & \\ \Xhline{1pt}
     &  Input & & & 127$\times$127  & 255$\times$255 & 3 \\
     & conv1 & 11$\times$11 & 2 & 59$\times$59 & 123$\times$123 & 96 \\
     C1 & pool1 & 3$\times$3 & 2 & 29$\times$29 & 61$\times$61 & 96 \\
     & conv2 & 5$\times$5 & 1 & 25$\times$25 & 57$\times$57 & 256 \\
     C2 & pool2 & 3$\times$3 & 2 & 12$\times$12 & 28$\times$28 & 256 \\
     C3 & conv3 & 3$\times$3 & 1 & 10$\times$10 & 26$\times$26 & 192 \\
     C4 & conv4 & 3$\times$3 & 1 & 8$\times$8 & 24$\times$24 & 192 \\
     C5 & conv5 & 3$\times$3 & 1 & 6$\times$6 & 22$\times$22 & 128 \\\Xhline{1pt}
    \end{tabular}
}
%\end{center}
\label{tb5}
%\vspace{-1em}
\end{table}

\begin{table}[t]
\caption{The Area Under the Curve (AUC) score for One-Pass Evaluation (OPE), Temporal Robustness Evaluation (TRE) and Spatial Robustness Evaluation (SRE), and speed (fps, * indicates GPU speed, otherwise CPU speed) on the OTB-50 dataset. The best results are shown in bold.}
\centering
\resizebox{0.75\linewidth}{!}{
\begin{tabular}{!{\vrule width1pt} c !{\vrule width1pt} c|c|c|c !{\vrule width1pt}}
    \Xhline{1pt}
    Method & LCT & SiamFC & SINT & EAST \\
     & \cite{MaYZY15} & \cite{Bertinetto2016} & \cite{TaoCVPR2016} & \\    \Xhline{1pt}
    AUC-OPE & 0.612 & 0.612 & 0.625 & \textbf{0.638} \\
    AUC-TRE & 0.594 & 0.621 & 0.643 & \textbf{0.662} \\
    AUC-SRE & 0.518 & 0.554 & 0.579 & \textbf{0.591} \\
    Speed & \textbf{27} & 86* & 4* & 23/\textbf{159}* \\ \Xhline{1pt}
    \end{tabular}
}
%\end{center}
\label{tb6}
%\vspace{-1em}
\end{table}

\begin{table}[t]
\caption{The accuracy $R_A$, robustness $R_R$ and average $R$ ranks under baseline and region noise experiments in VOT-14. $R_o$ is the overall (averaged) ranking for both experiments, which is used to rank the 38 trackers in the main paper. Our CPU/GPU speeds are reported in fps, while the speeds for the top 3 trackers are in EFO units, which roughly correspond to fps (\eg~the speed of the NCC baseline is 140 fps and 160 EFO).}
\centering
\resizebox{1.0\linewidth}{!}{
\begin{tabular}{!{\vrule width1pt} c !{\vrule width1pt} c|c|c !{\vrule width1pt} c|c|c !{\vrule width1pt} c|c !{\vrule width1pt}}
    \Xhline{1pt}
     & \multicolumn{3}{c!{\vrule width1pt}}{baseline} & \multicolumn{3}{c!{\vrule width1pt}}{region noise} & \multicolumn{2}{c!{\vrule width1pt}}{} \\ \cline{2-9}
     \raisebox{1ex}[0pt]{Tracker} & $R_A$ & $R_R$ & $R$ & $R_A$ & $R_R$ & $R$ & $R_{o}$ & Speed \\ \Xhline{1pt}
     EAST & \textbf{4.95} & \textbf{5.42} & \textbf{5.19} & \textbf{5.11} & \textbf{4.73} & \textbf{4.92} & \textbf{5.06} & \textbf{22/155} \\
     DSST & 5.41 & 11.93 & 8.67 & 5.40 & 12.33 & 8.86 & 8.77 & 7.66 \\
     SAMF & 5.30 & 13.55 & 9.43 & 5.24 & 12.30 & 8.77 & 9.10 & 1.69 \\
     KCF & 5.05 & 14.60 & 9.82 & 5.17 & 12.49 & 8.83 & 9.33 & 24.23 \\
     \Xhline{1pt}
    \end{tabular}
}
%\end{center}
\label{tb7}
\vspace{-1em}
\end{table}

\subsection*{Template Update}

It is worth noting that, in our feature cascade we explore the pixel and HOG layers before expensive deep layers. When processing the cheap pixel and HOG layers, we make use of fast correlation filters~\cite{henriques2015}. A correlation filter $w$ with the same size of image patch $x$ is learned by solving the Ridge Regression loss function
\begin{equation}
\label{eq8}
\min_{w} \sum_i \left( w^T x_i -y_i\right)^2 + \lambda \| w \|^2,
\end{equation}
where $y_i$ is the target response value, and $\lambda$ is the regularization parameter.

Solving this loss function is fast due to the efficient use of all shifted patches $x_i$ by exploiting the discrete Fourier transform. Besides fast speed, the correlation filter has another benefit of updating the template $w$ over time. However, this adaptive merit is not preserved for deep layers. Recall that for the deep convolutional layers $C1-C5$, we follow SiamFC~\cite{Bertinetto2016} to compute the similarity of a template image $z$ to all translated regions in search image $x$ by
\begin{equation}
\label{eq9}
F_l(z,x) = \varphi_l(z) \ast \varphi_l(x)+v\mathbb{I},
\end{equation}
where $\varphi_l$ is the convolutional feature embedding at layer $l$, and $v\in \mathbb{R}$ is an offset value.

Here $\varphi_l(z)$ can be treated as the convolutional template to compute the target responses, but is fixed to $\varphi_l(z_{t=1})$ from the first frame and is never updated during tracking. Then a question naturally arises: can we improve the performance by updating the template for deep layers?

\begin{figure}[t]
\begin{center}
\includegraphics[width=0.8\linewidth]{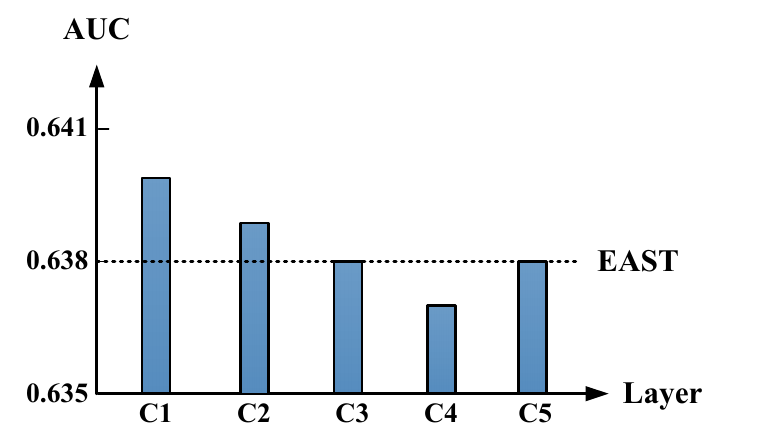}
\end{center}
\caption{The Area Under the Curve (AUC) score for One-Pass Evaluation (OPE) of template update for deep layers $C1-C5$.}
\label{fig7}
\end{figure}

To this end, we conduct the following experiment on OTB-50: we simply update $\varphi_l(z_t)$ as $\varphi_l(z_{t-1})$ from the previous frame, and record the accuracy if we separately update the convolutional layer $l$ from $C1-C5$. Fig.\ref{fig7} shows the AUC score when we update the template for each deep layer. Marginal gains are obtained on lower layers $C1-C2$, suggesting that they are less invariant and so would need to be updated more often. On the other hand, updating the top layer $C5$ leads to no difference, which is actually in line with the observations by SiamFC~\cite{Bertinetto2016} that always uses this invariant top layer for tracking. In the future, we can consider how to {\em learn} to update template online rather than just use the previous frame. Another promising direction is to further speedup the deep convolutional process by adopting the Fourier transform techniques.

\end{document}